\documentclass{bmvc2k}
\usepackage{multirow}
\usepackage{amsfonts}
\usepackage{hyperref}
\title{Re-Attention Transformer for Weakly Supervised Object Localization}

\addauthor{Hui Su}{suhui@zhejianglab.com}{1}
\addauthor{Yue Ye}{yeyue@zhejianglab.com}{1}
\addauthor{Zhiwei Chen}{chenzw@zhejianglab.com}{1}
\addauthor{Mingli Song}{brooksong@zju.edu.cn}{2}
\addauthor{Lechao Cheng}{chenglc@zhejianglab.com}{1*}

\addinstitution{
Zhejiang Lab,\\
Hangzhou, China
}
\addinstitution{
Zhejiang University,\\
Hangzhou, China
}

\def\etal{\emph{et al}\bmvaOneDot}

\runninghead{Hui Su, \etal }{Re-Attention Transformer for WSOL}

\begin{document}

\maketitle

\begin{abstract}
 Weakly supervised object localization is a challenging task which aims to localize objects with coarse annotations such as image categories. Existing deep network approaches are mainly based on class activation map, which focuses on highlighting discriminative local region while ignoring the full object. In addition, the emerging transformer-based techniques constantly put a lot of emphasis on the backdrop that impedes the ability to identify complete objects. To address these issues, we present a re-attention mechanism termed token refinement transformer (TRT) that captures the object-level semantics to guide the localization well. Specifically, TRT introduces a novel module named token priority scoring module (TPSM) to suppress the effects of background noise while focusing on the target object. Then, we incorporate the class activation map as the semantically aware input to restrain the attention map to the target object. Extensive experiments on two benchmarks showcase the superiority of our proposed method against existing methods with image category annotations. The source code is available in \url{https://github.com/su-hui-zz/ReAttentionTransformer}.
 \end{abstract}
\section{Introduction}

Recently, there emerges a surge of interests in localizing activation map with weakly supervision, for example with only image-level annotations, arising in different downstream applications. Weakly supervised learning has received a lot of attention since it minimizes the demand for fine-grained annotations so as to involve less human labeling effort. This work attempts to localize objects with only image category supervision, also termed weakly supervised object localization (WSOL).

\begin{figure}[t]
\centering
\includegraphics[scale=0.45]{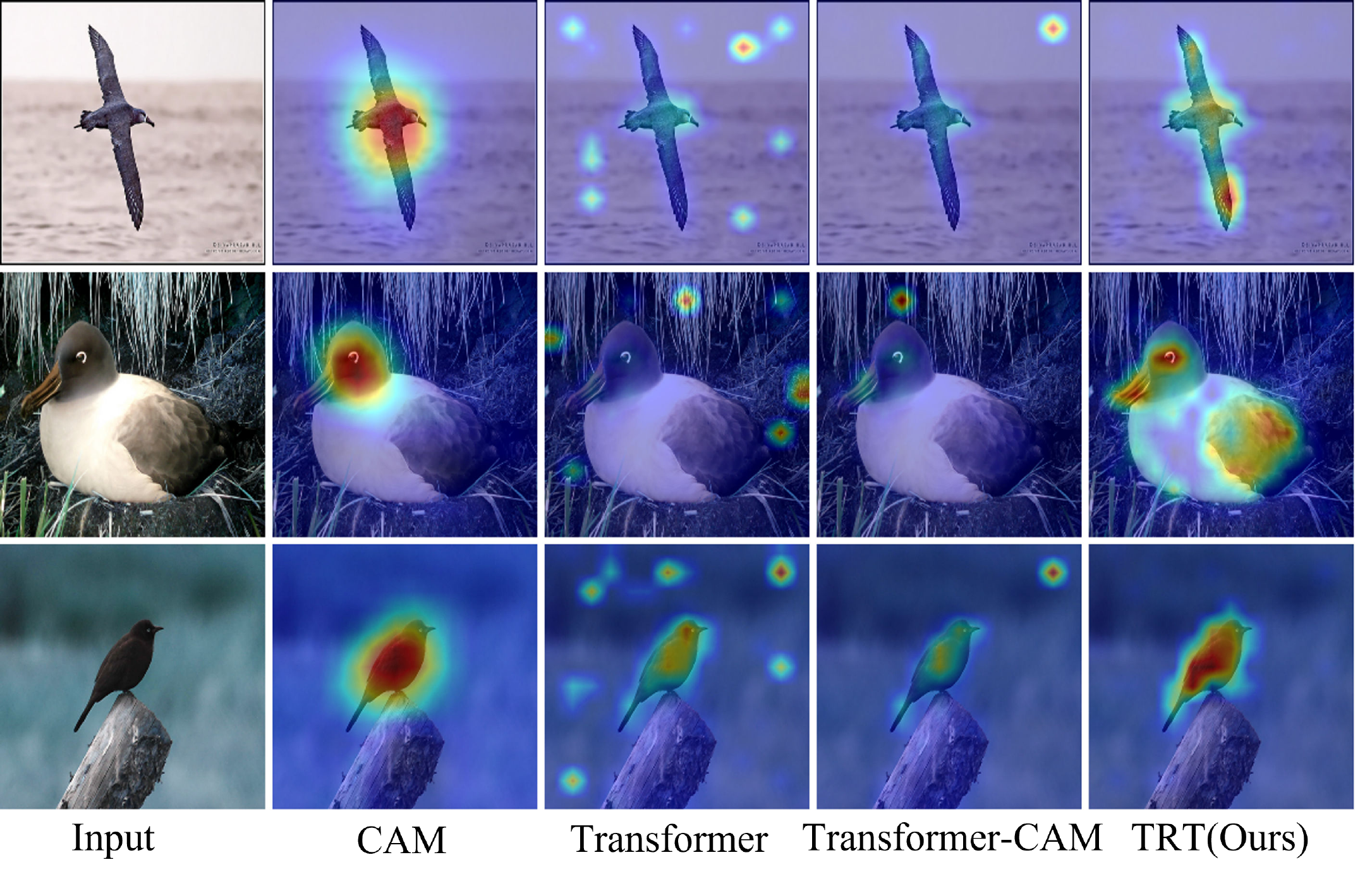}
\vspace{0.5em}
\caption{The comparisons between our proposed token refinement transformer (TRT) and existing approaches.  } \label{fig:teaser} 
\vspace{-2em}
\end{figure}

The dramatic progress of deep neural networks (DNNs) also provide tremendous benefits with impressive gains in the task of WSOL. The cornerstone work is the Class Activation Mapping(CAM)\cite{zhou2016learning} that enables classification neural networks to highlight the part of interest by computing a weighted sum of the last convolutional feature maps. However, CAM focuses only on the most discriminative features instead of the entire object areas due to its locality characteristic\cite{lee2021anti, ahn2018learning, xue2019danet, bae2020rethinking}. To alleviate this issue, much efforts have been made, such as adversarial erasing\cite{zhang2018adversarial, mai2020erasing, wei2017object}, multi-task joint training\cite{shen2019cyclic}, novel networks\cite{choe2019attention, zhang2020inter} and divergent activation\cite{singh2017hide, xue2019danet}. Most of these CAM-based approaches first compute local features and proceed to acquire adjacent object pieces to embrace the whole target. However, the underlying limitations of CAM still exist despite the fact that these approaches do encourage a better activation region. Another popular alternative is to leverage the long-range dependencies from vision transformer~\cite{dosovitskiy2020image} to find the object location. But this kind of approaches always suffer from irrelevant background for the reasons that canonical transformer-based methods generate patch tokens by splitting an image into a series of ordered patches and compute global relations in each layer. Therefore Gao et al. propose TS-CAM\cite{gao2021ts} by integrating both transformer and CAM to mitigate those drawbacks. As illustrated in Figure~\ref{fig:teaser}, CAM accentuates the most discriminative local region while transformer distracts attention from target to the background area. The results in penultimate column showcase that current ensemble approach can not deal with those issues even along with promising performance.

In this paper, we propose a re-attention strategy based on token refinement transformer (TRT) to grasp objects of interest more precisely. Specifically, TRT introduces a novel module named token priority scoring module (TPSM) to suppress the effects of background noise in transformer. We proceed to incorporate the class activation map as the semantically aware guidance to restrain the attention map to the target object. The TPSM attempts to re-score the important regions of strong response obtained by attention map in transformer to minimize the impact of jumbled background as much as possible. In addition, we also reveal that adaptive thresholding based on sampling over cumulative distribution function is superior to the regular thresholding or topK strategy. The contributions are summarized as follows: 
\begin{itemize}
\item we propose a re-attention mechanism termed token refinement transformer (TRT) which highlights the precise object of interest.
\item we propose an adaptive thresholding strategy based on sampling over cumulative importance that improves the performance significantly in the task of WSOL.
\item The experimental results show convincing results of both qualitative and quantitative compared to existing approaches on ILSVRC\cite{russakovsky2015imagenet} and CUB-200-2011\cite{wah2011caltech}.
\end{itemize}

\section{Related Work}
\subsection{CAM-based Attention Map Generation}
The milestone work of the class activation map (CAM~\cite{zhou2016learning}) aims to localize the discriminative regions for a specific category. In this study, we grouped CAM techniques into two categories: activation-based approaches~\cite{zhou2016learning,wang2020score,wang2020ss,naidu2020cam,ramaswamy2020ablation} and gradient-based approaches~\cite{selvaraju2017grad,chattopadhay2018grad,omeiza2019smooth,fu2020axiom,jiang2021layercam}. Typical activation-based works dispose of the reliance on the gradient as they only require access to class activation map and the channel-wise confidence. For example, Wang et al.~\cite{wang2020score} propose Score-CAM that computes the importance weights with the channel-wise increase of confidence performed on the activation map. Further, an extended version named SS-CAM~\cite{wang2020ss} provides better post-hoc explanations about the centralized target object in the image using smoothing to obtain a better understanding. In addition, other popular works, such as integrated score CAM (IS-CAM)~\cite{naidu2020cam}, attempt to alleviate the issue of saturation and false confidence and also achieve significant results. 
Apart from activation-based approaches, researchers often employ gradient information to enhance the salience of the image. Gradient-based methods highlight the important regions based on the back-propagated gradients of the specific class. However, the naive Grad-CAM~\cite{selvaraju2017grad} often fails to localize objects in an image when multiple occurrences of the same category exist. Chattopadhyay et al. propose Grad-CAM++~\cite{chattopadhay2018grad} to overcome above issue by introducing pixel-wise weighting of the gradients of the output with respect to a particular region in the final feature map. Moreover, the work of smooth Grad-CAM++~\cite{omeiza2019smooth} combines both smoothing and Grad-CAM++~\cite{selvaraju2017grad} to enable either visualizing a layer, subset of feature maps, or subset of neurons within a feature map.
Many additional techniques, including XGrad-CAM~\cite{fu2020axiom}, LayerCAM~\cite{jiang2021layercam} and others, have also contributed to the field's progress and delivered favorable results.

\subsection{Transformer-based Attention Map Generation}
Vision transformer~\cite{dosovitskiy2020image,hao2022attention,zhang2022long,fang2022cross} has attracted increasing attentions in recent studies, for example image classification~\cite{graham2021levit,liu2021swin,touvron2021training,wang2021pyramid,yuan2021tokens}, object detection~\cite{carion2020end,dai2021up,sun2021rethinking} and semantic segmentation~\cite{xie2021segformer}. However, vision transformers generate tokens by splitting an image into a series of ordered patches and computing global relations between tokens which inevitably introduces background noise. Existing works have made earlier attempts to explore the importance over tokens on fine-grained datasets~\cite{he2021transfg,wang2021feature} to alleviate the global issue. He et al.~\cite{he2021transfg} propose a part selection module that integrate all attention weights of the transformer into an attention map for guiding the network to form discriminative tokens. Inspired by similar motivation that the classification token in the transformer layer pays more attention to the global information which yields poor explanations, Wang et al.~\cite{wang2021feature} propose a feature fusion transformer to aggregate the dominant tokens from each transformer layer to compensate for the local information. 
In addition, few studies have been proposed to facet the highlighting region~\cite{chen2022lctr,gao2021ts} based on both CAM and transformer. Despite the fact that solid numerical results have been achieved, the highlight regions still bring issues derived from both CAM and Transformer. 
\subsection{Weakly Supervised Object Localization}
Conventional works, such as \cite{li2016weakly,cinbis2016weakly} formulate the task of weakly supervised object localization (WSOL) as a multiple instance learning (MIL) problem. However the canonical approaches to the problem of MIL are typically initialization-sensitive which are prone to get stuck in local minimal with high generalization error. Thus many approaches~\cite{kumar2010self,deselaers2010localizing,cabral2014matrix, cinbis2016weakly,tang2017multiple, zhang2021weakly} attempt to obtain good initial values or to explore the refinement methods in post-processing. Recent there have been sufficient advancements in deep neural networks (DNNs) to cope with the challenging WSOL task, such as CAM-based~\cite{zhou2016learning,selvaraju2017grad} and transformer-based approaches~\cite{dosovitskiy2020image,graham2021levit,gao2021ts}. While in this work, we follow the trend to deal with the problems caused by global background noise and local discriminative regions with the proposed re-attention mechanism.

\section{Method}
In this section, we first revisit the preliminaries for vision transformer~\cite{dosovitskiy2020image}. Then we discuss the details of the proposed token refinement transformer (TRT) for WSOL.

\begin{figure}[h]
\centering
\includegraphics[width=0.95\linewidth]{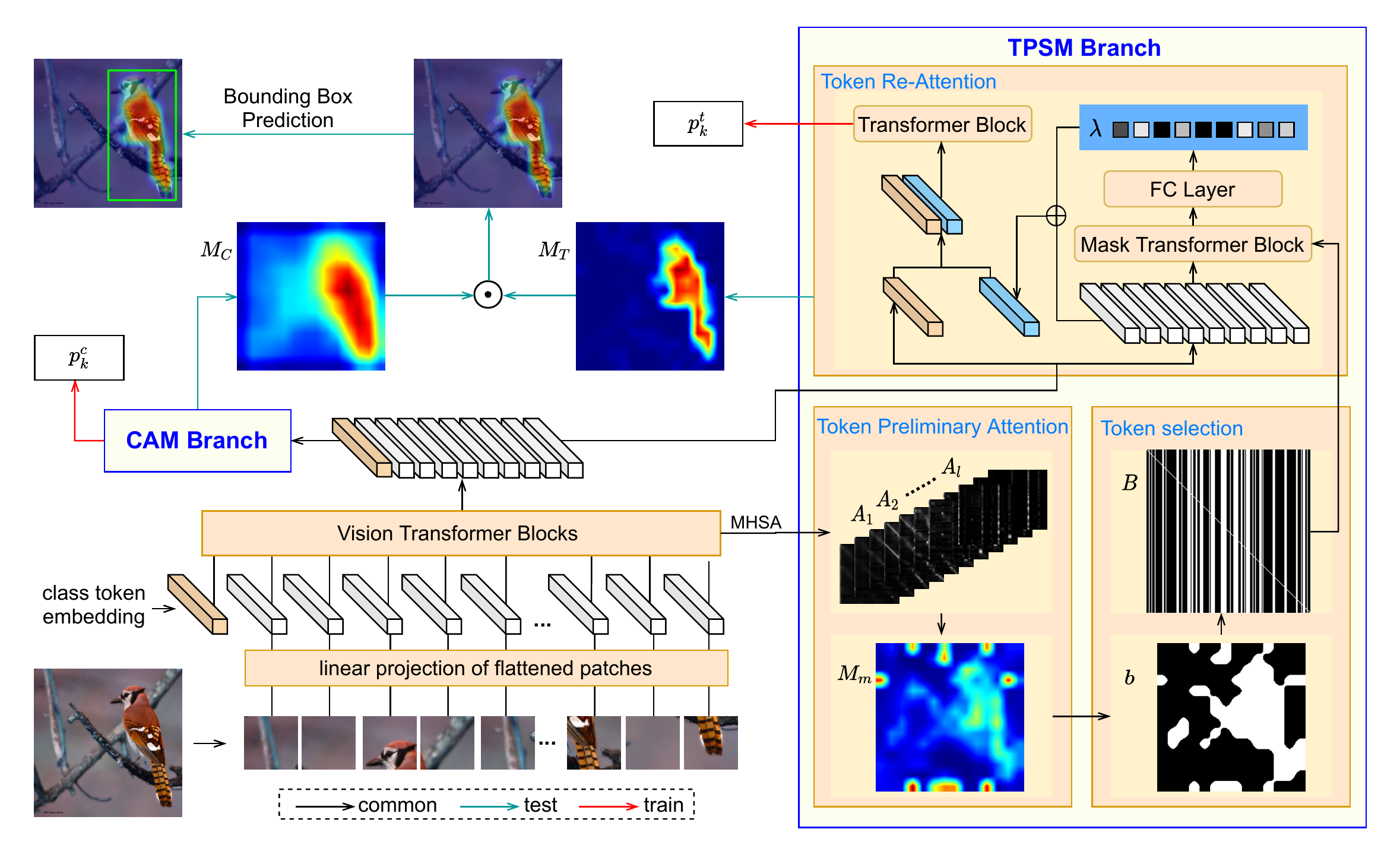}
\caption{Token Refinement Transformer (TRT) framework. TRT consists of two branches, Token Priority Scoring Module (TPSM) and CAM, respectively. TPSM attempts to generate context-aware features $\mathbf{M}_{T}$ that contribute most to the target class. CAM is introduced to obtain discriminative features $\mathbf{M}_{C}$. We finally get the attention map $\mathbf{M}$ by performing element-wise multiplication as $\mathbf{M} = \mathbf{M}_{C} \odot \mathbf{M}_{T} $.   } \label{fig:framework} 
\end{figure}

\subsection{Preliminary}
Let us consider $I \in \mathbb{R}^{W\times H \times 3}$ as an input image, we split the image $I$ based on the patch size $P$, resulting in N ( $N = [\frac{W}{P}] \times [\frac{H}{P}]$) non-overlap flattened patch blocks $\mathbf{x}_p \in \mathbb{R}^{N\times (3\times P^2)}$. Each patch block $\mathbf{x}^n_p (n\in \{1,...,N\})$ is linearly projected into $D$ dimensional patch embedding before being fed into transformer block. As part of the embeddings, an extra learnable class token $\mathbf{z}_0^{cls}\in\mathbb{R}^{1\times D}$ is introduced. In addition, we incorporate a position embedding $\mathbf{E}_{pos} \in \mathbb{R}^{(N+1)\times D}$ to form the whole patch embedding for transformer as follows:
\begin{equation}\label{eq1}
    \mathbf{Z}_0 = [\mathbf{z}_0^{cls};E(\mathbf{x}_p^1);E(\mathbf{x}_p^2);...;E(\mathbf{x}_p^N)] + \mathbf{E}_{pos}
\end{equation}
Where $E(.)$ indicates patch embedding projection and $\mathbf{Z}_0\in\mathbb{R}^{(N+1)\times D}$ is the input for the first transformer block.

\subsection{Overview}

We define $\mathbf{Z}_{l}\in\mathbb{R}^{(N+1)\times D}~(l \in \{1, ..., L\})$ as the output feature embedding for $l$-th transformer block. As illustrated in Figure~\ref{fig:framework}, the output of the penultimate transformer $\mathbf{Z}_{L-1}$ is fed into two branch, one of which is Token Priority Scoring Module (TPSM) that aims to re-attention patch tokens with the proposed adaptive thresholding strategy while the other branch compute standard class activation map.

In training stage, we evaluate the inconsistency between the output and ground truth in both two branches with commonly used cross entropy loss. Let $\mathbf{z}^{c}_{L-1} \in \mathbb{R}^{1\times D}$ and $\mathbf{z}^{p}_{L-1} \in \mathbb{R}^{N \times D}$ be the output class token and patch token at the penultimate layer, respectively. For CAM branch, we reshape $\mathbf{z}^{p}_{L-1}$ to $\mathbf{z'}^{p}_{L-1} \in \mathbb{R}^{\sqrt{N}\times\sqrt{N} \times D}$, which serves as the effective input for the subsequent convolution layer. Further, the output feature is globally average pooled followed by a softmax layer to get classification prediction $\mathbf{p}^c$. 
For TPSM branch, $\mathbf{Z}_{L-1}(=[\mathbf{z}^{c}_{L-1};\mathbf{z}^{p}_{L-1}])$ is subsequent processed with the re-attention module to obtain classification probability $\mathbf{p}^t$. The loss function is thus defined as:

\begin{equation}\label{eq2}
    L_{ce} = - \sum_{k=1}^{K}{y_k ( \log{p^c_k}}  + \log{p^t_k})
\end{equation}
where $L_{ce}$ denotes cross entropy loss function. $K$ is the number of category. $y_k$ is the binary indicator (0 or 1) if class label $k$ is the correct classification for the observation. $p^c_k$ and $p^t_k$ denote the output probability for class $k$ in CAM and TPSM branch, respectively.

During testing, we first forward $\mathbf{z}^{p}_{L-1}$ into TPSM to obtain context-aware feature maps $\mathbf{M}_T \in \mathbb{R}^{\sqrt{N} \times \sqrt{N}}$ by performing a re-attention operation on patch tokens. More details will be discussed in Section~\ref{sec:tpsm}. For the counterpart, we apply standard CAM to generate class specific activation maps $\mathbf{M}_C \in \mathbb{R}^{K \times \sqrt{N} \times \sqrt{N}}$. Consequently, attention maps can be obtained by:
\begin{equation}\label{eq3}
    \mathbf{M} = \mathbf{M}_T \odot\mathbf{M}_C
\end{equation}
where $\odot$ denotes an element-wise multiplication operation. The proposed context-aware feature map $\mathbf{M}_T \in \mathbb{R}^{\sqrt{N} \times \sqrt{N}}$ is class-agnostic so we incorporate the class-aware activation map $\mathbf{M}_C \in \mathbb{R}^{K \times \sqrt{N} \times \sqrt{N}}$ as the semantically guidance to restrain the attention map to the target class. We further resize the maps $\mathbf{M}$ to the same size of the original images by bilinear interpolation. Specifically, we separate the foreground from the background using a defined threshold as described in~\cite{zhou2016learning}. Then we look for the tight bounding boxes that enclose the most linked region in the foreground pixels. Finally with the grid search approach, the thresholds for obtaining bounding boxes are updated to the optimal values.

\subsection{Token Priority Scoring Module}\label{sec:tpsm}


Our token priority scoring module(TPSM) consists of three components as indicated in Figure~\ref{fig:framework}. Firstly, we generate a preliminary attention map by exploiting long-range dependencies of class token and patch tokens over transformer blocks. Then an adaptive thresholding strategy is introduced to screen out patch tokens with high response in preliminary attention map. Finally, we perform re-attention operation on the selected tokens to capture more effective global relationships. We will detail the descriptions in following three subsections.



\subsubsection{Token Preliminary Attention}

Multi-head self-attention(MHSA)~\cite{vaswani2017attention} is widely used in transformer to model the long range dependency. We first compute the preliminary self-attention for $l$-th transformer block as:
\begin{equation}
    \mathbf{A}_l = softmax(\frac{\mathbf{Q}_l \mathbf{K}_l^{\mathbf{T}}}{\sqrt{D}})
\end{equation}

where $\mathbf{Q}_l$ and $\mathbf{K}_l$ are the query and key representations projected from the previously output $\mathbf{Z}_{l-1}$. $D$ indicates dimension of patch embeddings. $\mathbf{T}$ is a transpose operator. 

We later investigate the characteristic of  $\mathbf{A}_l \in\mathbb{R}^{(1+N)\times (1+N)}$. It is found that attention vector in the first row, which records dependency of class token to patch tokens, is driven to highlight object regions when Eq.\ref{eq2} is optimized. We aggregate these attention vectors over transformer blocks as $\mathbf{m} = \sum_{l=1}^{L-1}\mathbf{A}_l[0,1:]$. $\mathbf{m} \in \mathbb{R}^{1\times N}$ is then reshaped back to $\mathbf{M}_m \in \mathbb{R}^{\sqrt{N} \times \sqrt{N}}$ as a preliminary attention map as illustrated in Figure~\ref{fig:framework}.

\subsubsection{Token Selection Strategy}
Token preliminary attention considers the cumulative dependency between patch tokens and class tokens based on the attention map of multi-head self-attention. As shown in Figure~\ref{fig:framework}, we are suggested to suppress the irreverent background response to highlight the objective regions. Intuitively, we may pick up the top $k$ largest responses or responses that exceed a fixed threshold $\tau$. However we experimentally find that these two basic strategies do not work well in practice. Thus we propose an adaptive thresholding strategy based on sampling over cumulative importance that boosts the performance significantly.

We first calculate the cumulative distribution function $F$ of $\mathbf{m}$ and define strictly monotone transformation $\mathbb{T}: U \sim [0,1] \mapsto \mathbb{R}$ as the inverse function, thus
\begin{equation}
    F(x) = \mathbf{P}_r(\mathbf{m} < x) =  \mathbf{P}_r( \mathbb{T}(U) <x) = \mathbf{P}_r( U < \mathbb{T}^{-1}(x)) = \mathbb{T}^{-1}(x)
\end{equation}
$\mathbf{P}_r$ is the probability function. $F$ is considered as the inverse function of $\mathbb{T}$, or $ \mathbb{T}(u) = F^{-1}(u), u\sim [0,1]$. Specifically, we first sort the values in $\mathbf{m}$ from high to low and calculate the cumulative attention. Then the adaptive threshold $\tau'$ is obtained based on the inverse transform sampling given the contribution $u$ over cumulative attention. More theoretical details about adaptive threshold generation can be referred to "Inverse Transform Sampling." $u$ controls the proportion of selected token attention in token preliminary attention. During the training and inference stages, $\tau'$ varies adaptively for different images and can achieve good results for objects with different scales. Therefore, we are capable of generating adaptive thresh $\tau'$ from $ F^{-1}(u)$. We denote $\mathbf{b} = [\mathbf{m} > \tau'] $ as the binary mask for the existence of selected patch tokens.

\subsubsection{Token Re-Attention}
Let $\mathbf{I}_N$ be the identity matrix with size $N$. To achieve the goal of drawing more attention to class-specific objectiveness instead of background, we generate the selection matrix $\mathbf{B} \in \mathbb{R}^{N \times N}$ for token re-attention as:
\begin{equation}
    \mathbf{B} = \mathbf{J} \otimes \mathbf{b} + \mathbf{J} \otimes (\mathbf{J}^\mathbf{T} - \mathbf{b} ) \odot \mathbf{I}_N
\end{equation}
$\mathbf{J} \in \mathbb{R}^{N \times 1}$ is a matrix where every element is equal to one. $\otimes$ means tensor product. $\mathbf{B}$ is a binary matrix where each entry $\mathbf{B}_{i,j} $ means the $j$-th token will contribute to the update of the $i$-th token. We replace self-attention modules in transformer block by masked self-attention modules as follows:

\begin{equation}
    \mathbf{S} = \frac{\mathbf{Q}_{L-1} \mathbf{K}_{L-1}^{\mathbf{T}}}{\sqrt{D}}
\end{equation}
\begin{equation}
     \mathbf{A}^{r}_{ij} =  \frac{\exp(\mathbf{S}_{ij}) * \mathbf{B}_{ij}}{\sum_{k=1}^N \exp(\mathbf{S}_{ik}) * \mathbf{B}_{ik}}
\end{equation}

Patch tokens $\mathbf{z}^{p}_{L-1}$ are fed to mask transformer block followed by a fully connected layer and a mask softmax layer, resulting in importance weights $\mathbf{\lambda}$. For training stage, the fusion embedding is generated by computing a weighted sum of importance weights $\mathbf{\lambda}$ with patch tokens $\mathbf{z}^{p}_{L-1}$. We further concatenate class embedding and fusion embedding to feed into final transformer block to yield classification loss. For inference stage, we retrieve the weights from the original relation $\mathbf{m}$ for pruned tokens. Hence, the re-attention vector is then defined as:

\begin{equation}
    r = \frac{\sum_{k=1}^N{\mathbf{m}_k *  \mathbf{b}_k}}{\sum_{k=1}^N{\mathbf{\lambda}_{k}}}
\end{equation}
\begin{equation}
    \mathbf{m'} = \mathbf{m} \odot  (\mathbf{J}^{\mathbf{T}} - \mathbf{b}) + \mathbf{\lambda} * r
\end{equation}

We further reshape $\mathbf{m'} \in \mathbb{R}^{N}$ to yield context-aware feature map $\mathbf{M}_{\mathbf{T}} \in \mathbb{R}^{\sqrt{N} \times \sqrt{N}}$.

\section{Experiments}

\subsection{Experimental Settings}
Our method is implemented under PyTorch with four 32GB V100 GPUs. Adam is used for optimization, and the weight decay is set to $5\times10^{-4}$. The basic backbone used is the Deit-Base\cite{touvron2021training} pre-trained on ImageNet-1K\cite{russakovsky2015imagenet}. We evaluate our proposed framework on two commonly used benchmark datasets (ILSVRC\cite{russakovsky2015imagenet} and CUB-200-2011\cite{wah2011caltech}) in WSOL. For CUB-200-2011\cite{wah2011caltech}, each input image is resized to 256 and randomly cropped to 224. The batch size is 128, and the adaptive uniform threshold is 0.65. We train the backbone and TPSM branches concurrently for 30 epochs, then fix the backbone and TPSM branch parameters and train the CAM branch for additional 15 epochs. For ILSVRC\cite{russakovsky2015imagenet}, we follow \cite{touvron2021training} to employ AutoAugment~\cite{cubuk2019autoaugment} on the inputs. Similar to CUB-200-2011, we also train the backbone and TPSM branches in first 12 epochs and proceed to tune CAM branch for 10 epochs. $u$ is set to 0.95 for ILSVRC. * means we reproduce the results based on the provided codes unless otherwise stated. We adopt the evaluation metrics as  previous works~:
\begin{itemize}
\item Gt-Known \emph{Loc.Acc}\cite{zhang2018self} is the localization accuracy with ground-truth class that provides a positive value when Intersection over Union(IOU) between the predicted box and ground truch box is larger than 50\%.
\item Top-1/Top-5 \emph{Loc.Acc}\cite{russakovsky2015imagenet} means Top-1/Top-5 localization accuracy that fulfills: 1) the predicted class is correct. 2). the IOU between the predicted box and ground truch box is larger than 50\%.
\item MaxBoxAccV2 \cite{choe2022evaluation}averages the localization accuracy across IOU thresholds in [30\%, 50\%, 70\%], regardless of classification results.
\end{itemize}
\begin{table}[tp]
    \centering
        \begin{tabular}{c|c|c|c|c}
        \hline
        \multirow {2}{*}{Methods} & \multirow{2}{*}{Backbone} & \multicolumn{3}{c}{\emph{Loc.Acc}} \\ 
        \cline{3-5}
        ~ & ~ & Top-1 & Top-5 & Gt-Known \\
        \hline
        CAM\cite{zhou2016learning} & VGG16 & 44.2 & 52.2 & 56.0 \\
        SPG\cite{zhang2018self} & VGG16 & 48.9 & 57.2 & 58.9 \\
        SLT-Net\cite{guo2021strengthen} & VGG16 & 67.8 & - & 87.6 \\
        \hline
        CAM\cite{zhou2016learning} & InceptionV3 & 41.1 & 50.7 & 55.1 \\
        SPG\cite{zhang2018self} & InceptionV3 & 46.7 & 57.2 & - \\
        I2C\cite{zhang2020inter} & InceptionV3 & 66.0 & 68.3 & 72.6 \\
        SLT-Net\cite{guo2021strengthen} & InceptionV3 & 66.1 & - & 86.5 \\
        \hline
        TS-CAM\cite{gao2021ts} & Deit-B & 75.8 & 84.1 & 86.6 \\
        TS-CAM*\cite{gao2021ts} & Deit-B-384 & 77.8 & 88.6 & 90.8 \\
        \hline
        TRT(Ours) & Deit-B & 76.5 & 88.0 & 91.1 \\
        TRT(Ours) & Deit-B-384 & 80.5 & 91.7 & 94.1 \\
         \hline
        \end{tabular}
      
        \vspace{2.5mm}
        \caption{Experimental results on CUB-200-2011 for metrics of \emph{Loc.Acc}.}\label{tab:sota-cub-la}
\end{table}
\begin{table}[h]
    \centering
    \begin{tabular}{c|c|c}
    \hline
    Methods & Backbone & MaxBoxAccV2 \\
    \hline
    CAM\cite{zhou2016learning} & VGG & 63.70 \\
    ADL\cite{choe2019attention} & VGG & 66.30\\
    \hline
    VITOL\cite{gupta2022vitol} & Deit-B & 73.17 \\
    TS-CAM*\cite{gao2021ts} & Deit-B & 76.74 \\
    \hline
    TRT(Ours) & Deit-B & 82.08 \\
    \hline
    \end{tabular}
    \vspace{2.5mm}
    \caption{Experimental results on CUB-200-2011 for MaxBoxAccV2.} \label{tab:sota-cub-mbav2}
\end{table}
\subsection{Comparisons to the State-of-the-arts}
\begin{figure}[h]
\centering
\subfigure[CUB-200-2011] 
{
	\begin{minipage}[t]{0.48\linewidth}
	\centering     
	\includegraphics[width=\textwidth]{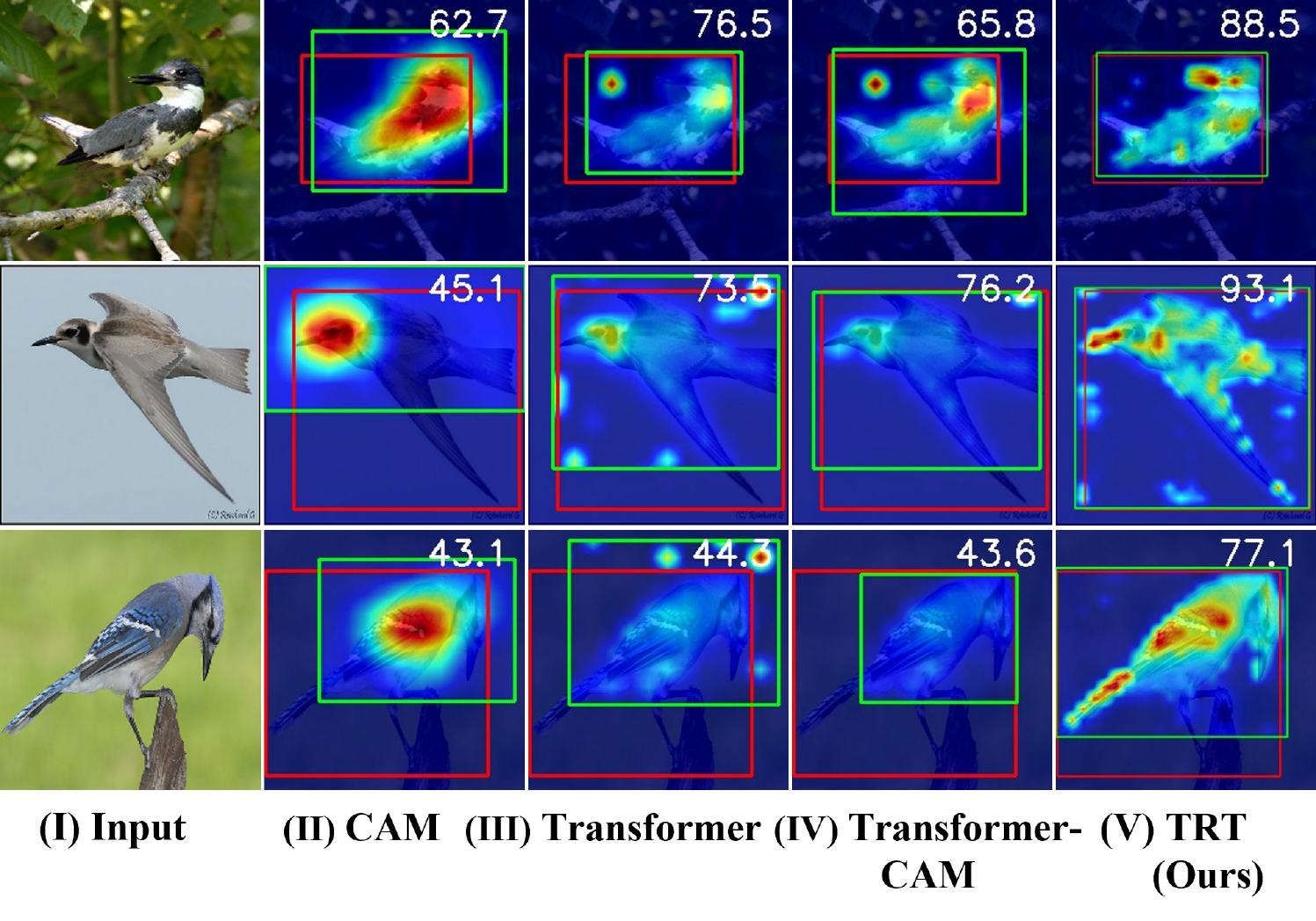}
	\end{minipage}
}
\subfigure[ILSVRC] 
{
    \begin{minipage}[t]{0.48\linewidth}
	\centering
	\includegraphics[clip,width=\textwidth]{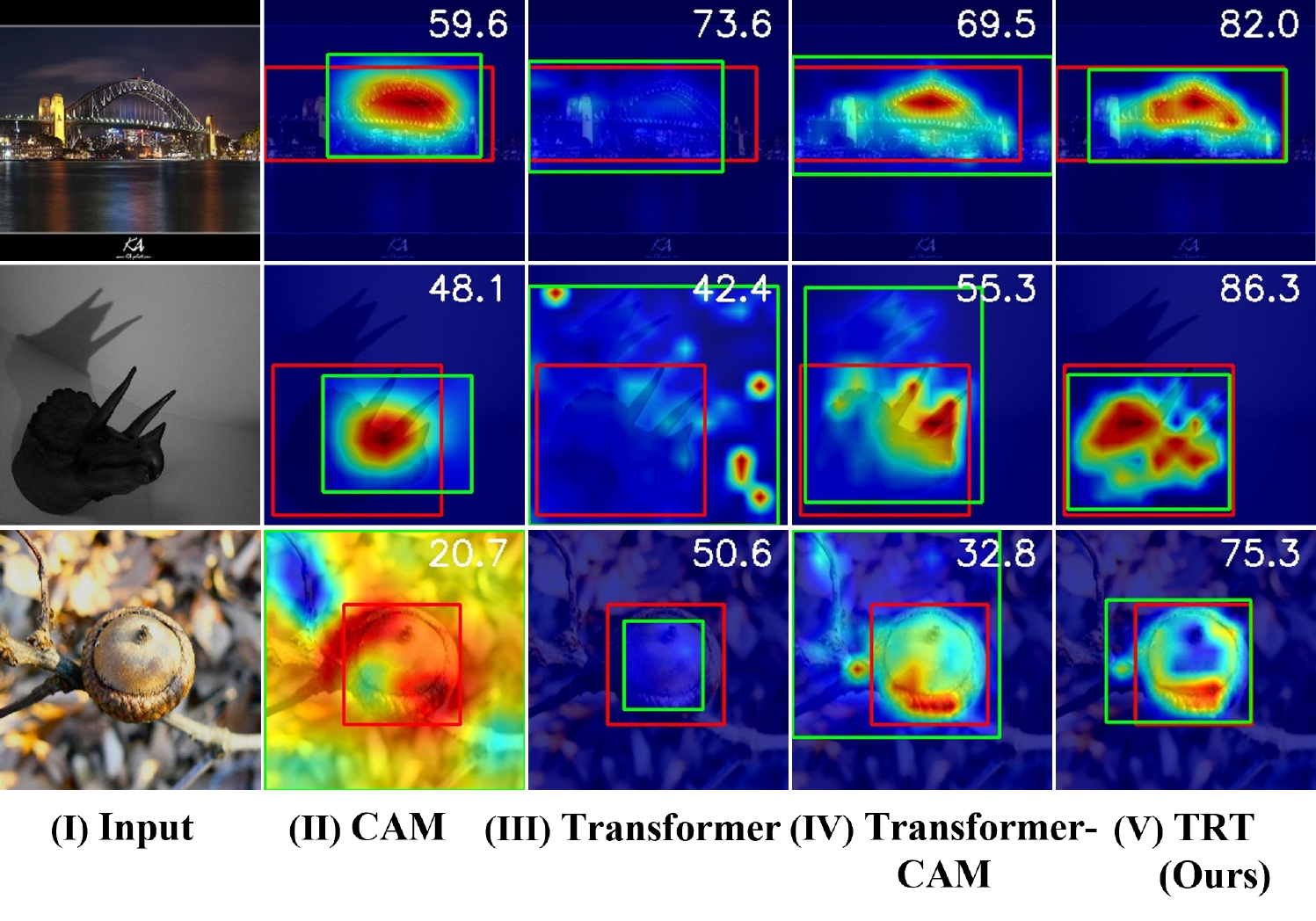}
	\end{minipage}
}
\caption{Visualization of localization maps on CUB-200-2011 and ILSVRC datasets. \textcolor{red}{Red} means ground truth and \textcolor{green}{green} means predicted bounding box.} 
\label{fig:performance} 
\end{figure}

Table \ref{tab:sota-cub-la} and Table \ref{tab:sota-cub-mbav2} showcase the competitive results of our proposed TRT framework on the CUB-200-2011. Overall, our approach achieves remarkable performance across all metrics compared to existing methods. Specifically, TRT outperforms state-of-the-art methods with a large margin in Gt-Known, yielding a localization accuracy of 91.1\% against 86.6\%. It can also be observed that the benefits over other approaches stand out more when higher quality images are used. As for ILSVRC, Table \ref{tab:sota-imagenet-la} demonstrated that TRT is superior to both existing CAM-based and transformer-based approaches. We have reproduced the results of \cite{chen2022lctr} and ~\cite{gao2021ts} on ILSVRC with backbones Deit-S and Deit-B based on the released codes. It shows that the results of \cite{chen2022lctr} and ~\cite{gao2021ts} based on Deit-B are even worse than Deit-S, although Deit-B has much more capacity. In contrast, our results on ILSVRC outperform \cite{chen2022lctr} and ~\cite{gao2021ts} on both Deit-S and Deit-B. And we obtain better performances based on Deit-B than those from Deit-S for our proposed approach. A reasonable explanation may be that inductive bias (the re-attention constraint, etc.) allows us to better use the model capacity in the vision transformer.
Figure~\ref{fig:performance} illustrates the qualitative results. When compared to CAM-based and Transformer-based techniques, our produced attention maps show visual superiority that focuses more on the objects.


\begin{table}[tp]
    \centering
        \begin{tabular}{c|c|c|c|c}
        \hline
        \multirow {2}{*}{Methods} & \multirow{2}{*}{Backbone} & \multicolumn{3}{c}{Loc.Acc} \\ 
        \cline{3-5}
        ~ & ~ & Top-1 & Top-5 & Gt-Known \\
        \hline
        CAM\cite{zhou2016learning} & VGG16 & 42.8 & 54.9 & 59.0 \\
        ADL\cite{choe2019attention} & VGG16 & 44.9 & - & - \\
        SLT-Net\cite{guo2021strengthen} & VGG16 & 51.2 & 62.4 & 67.2 \\
        \hline
        CAM\cite{zhou2016learning} & InceptionV3 & 46.3 & 58.2 & 62.7 \\
        ADL\cite{choe2019attention} & InceptionV3 & 48.7 & - & - \\
        SLT-Net\cite{guo2021strengthen} & InceptionV3 & 55.7 & 65.4 & 67.6 \\
        \hline
        TS-CAM*\cite{gao2021ts} & Deit-B & 47.8 & 60.0 & 64.4 \\
        LCTR*\cite{chen2022lctr} & Deit-B & 53.4 & 63.9 & 67.1 \\
        \hline
        TRT(ours)  & Deit-B & 58.8 & 68.3 & 70.7\\
        \hline
        \end{tabular}
        \vspace{2.5mm}
        \caption{Experimental results on ILSVRC for metrics of Loc.Acc.} \label{tab:sota-imagenet-la}
\end{table}

 \begin{figure}[h]
\centering
\subfigure[Token selection strategies] 
{
	\begin{minipage}[t]{0.48\linewidth}
	\centering     
	\includegraphics[width=\textwidth]{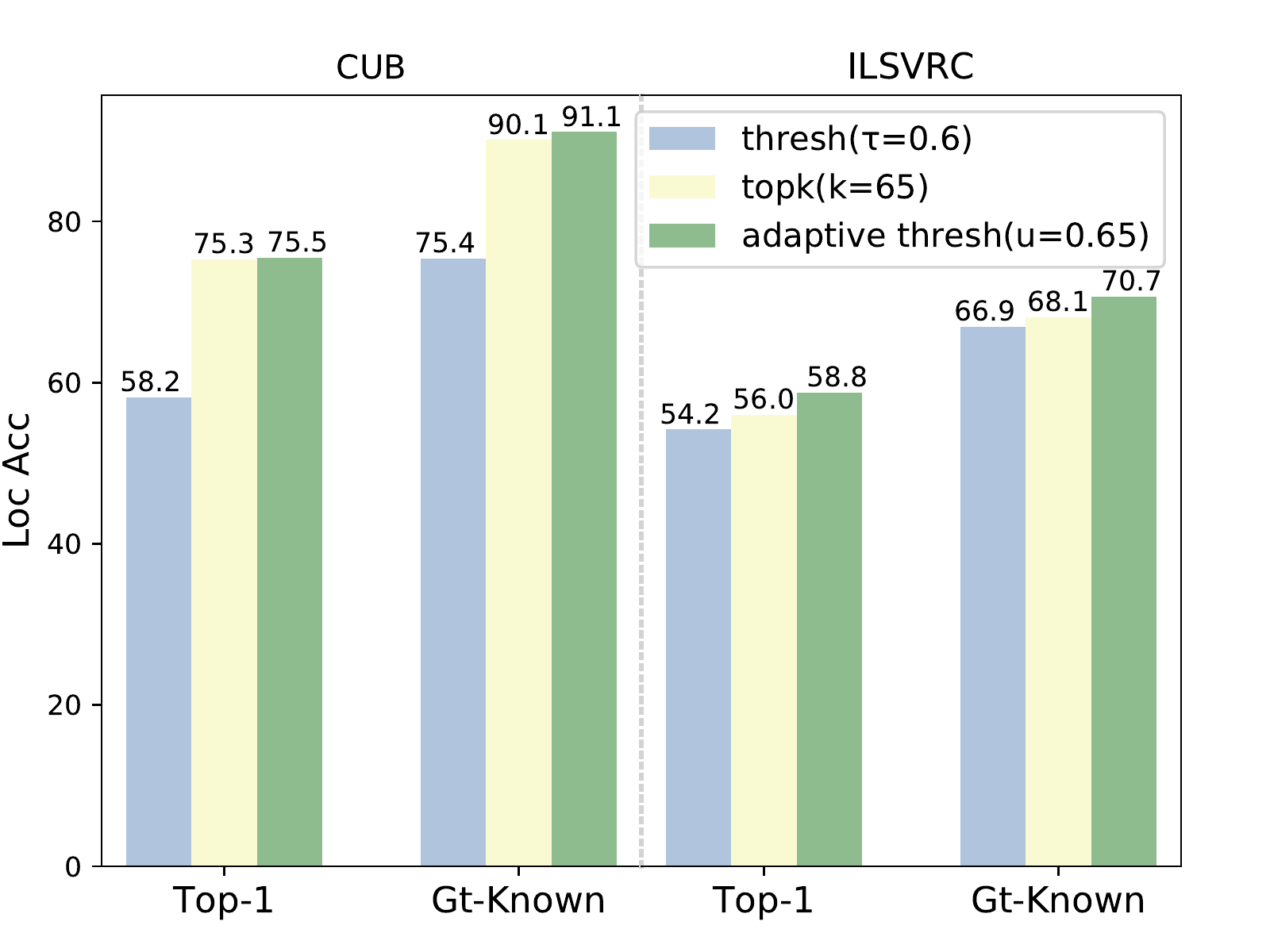}
	\end{minipage}
   
} \label{fig:tse}
\subfigure[The impact of uniform distribution] 
{
    \begin{minipage}[t]{0.48\linewidth}
	\centering
	\includegraphics[clip,width=0.91\textwidth]{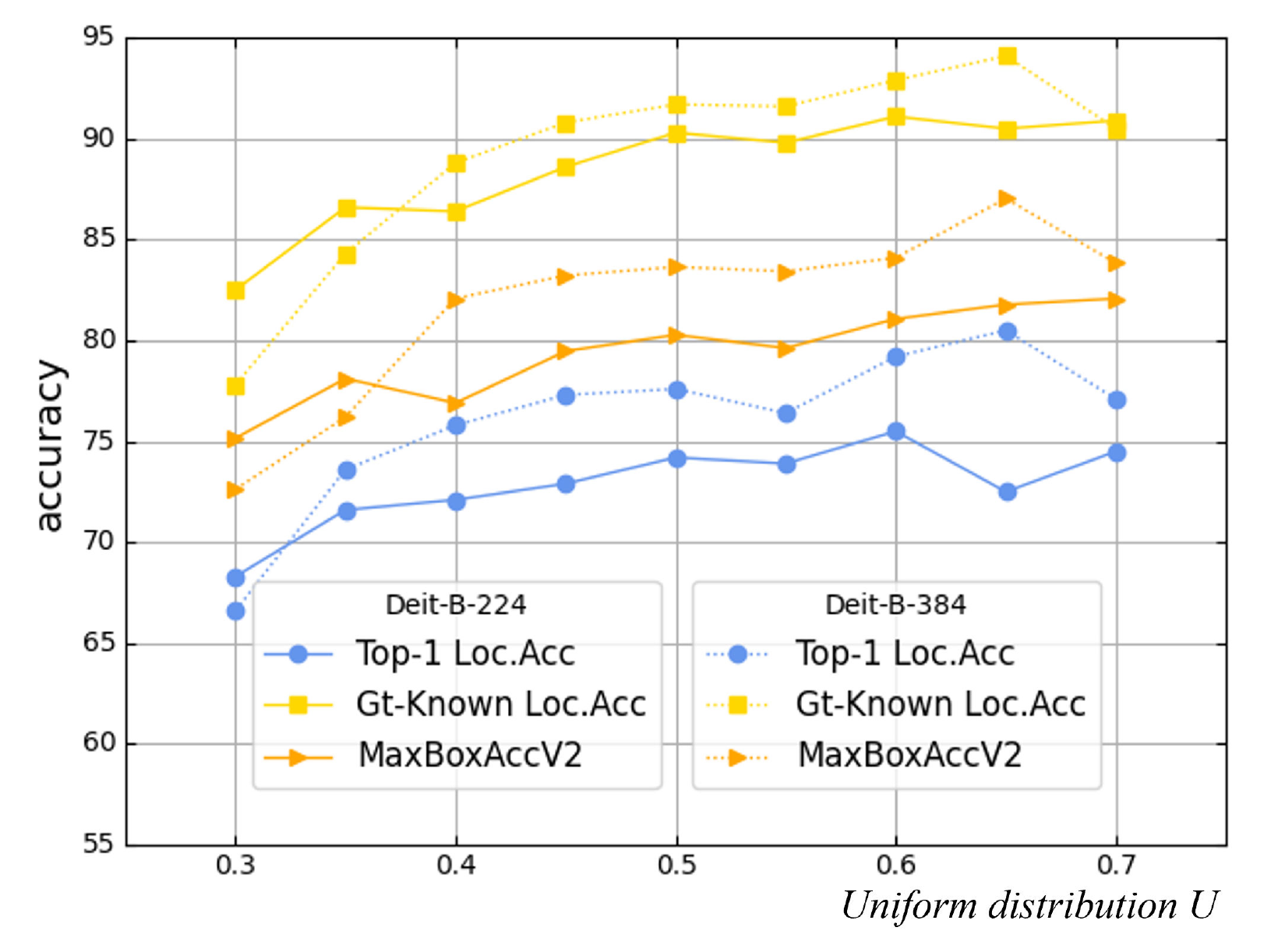}
	\end{minipage}
}\label{fig:ud}
\vspace{0.3mm}
\caption{Token selection module. } \label{fig:ablation-token-select}
\end{figure}

 \begin{figure}[h]
\centering
\subfigure[Quantitative results] 
{
	\begin{minipage}[t]{0.48\linewidth}
	\centering     
	\includegraphics[width=\textwidth]{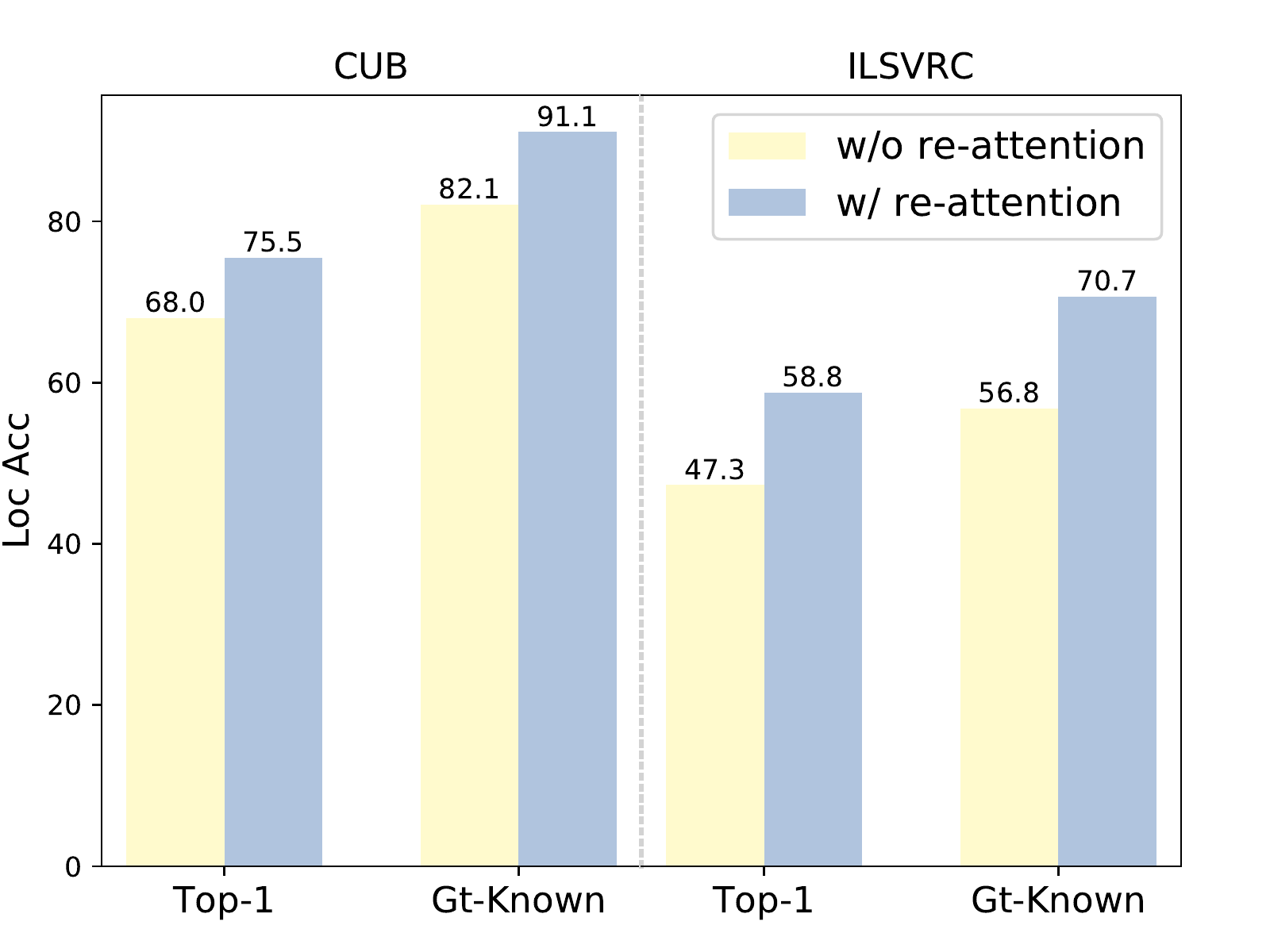}
	\end{minipage}
}
\subfigure[Qualitative results] 
{
    \begin{minipage}[t]{0.48\linewidth}
	\centering
	\includegraphics[clip,width=0.91\textwidth]{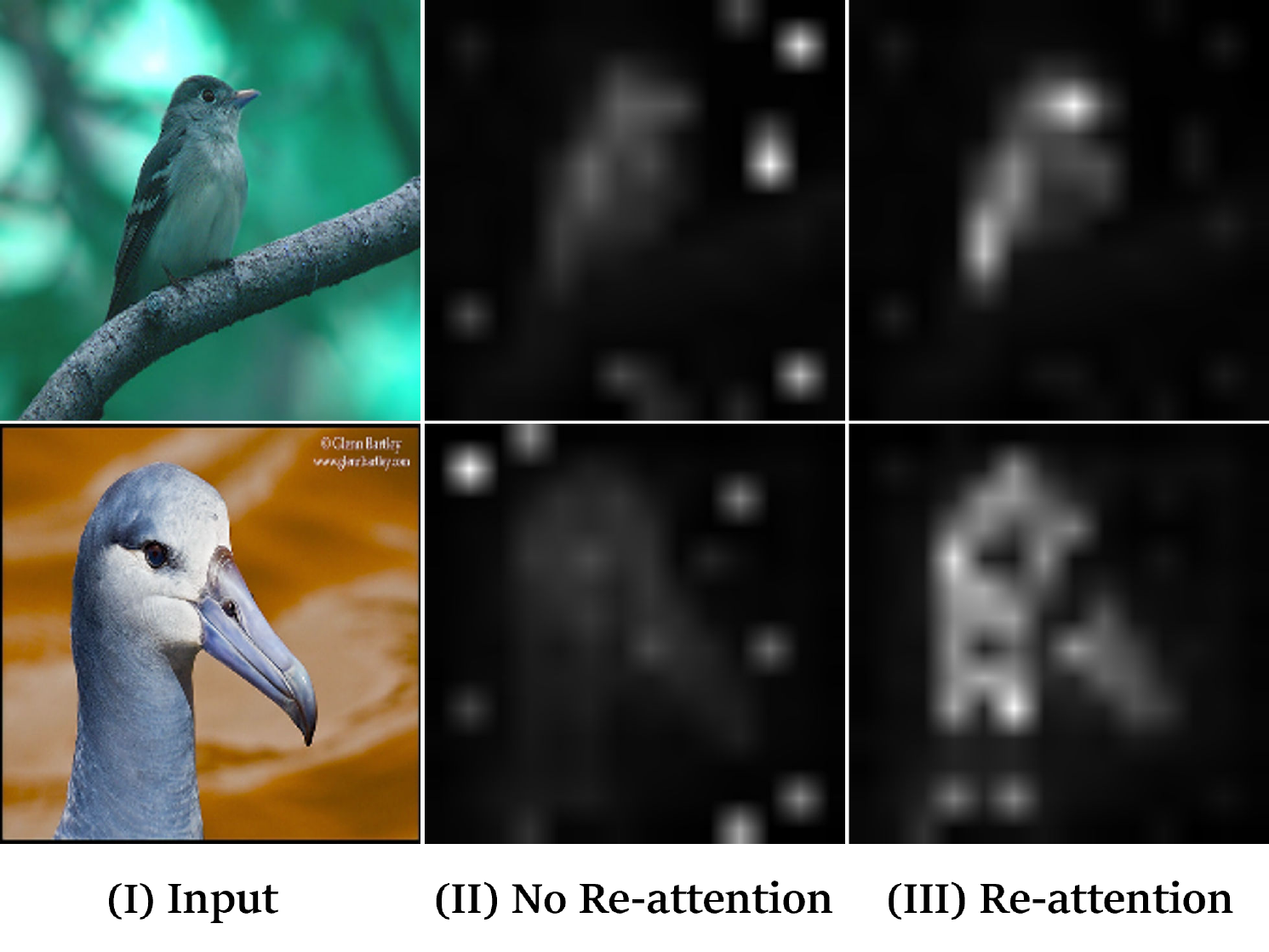}
	\end{minipage}
}
\vspace{0.3mm}
\caption{The impact of Re-Attention. }\label{fig:ram} %
\label{fig:ablation} 
\end{figure}

\subsection{Ablation Study}\label{sec:ablation}
In this section, we conduct ablation studies to verify the effectiveness of our proposed approach. First, we investigate the impact for different token selection strategies as shown in figure~\ref{fig:ablation-token-select} (a). The experimental results emphasize the necessity of the proposed adaptive thresholding method. Figure~\ref{fig:ablation-token-select} (b) empirically suggest us the default setting about u(=0.65) on CUB-200-2011.
In addition, we also did the ablations about the re-attention module. Both qualitative and quantitative results as shown in figure~\ref{fig:ram} reveal that the aforementioned re-attention block plays a critical role in framework.

\section{Conclusion}
In this work, we introduce a Re-Attention method called the token refinement transformer (TRT) that captures object-level semantics for the task of WSOL. To reduce the impact of background noise while concentrating on the target object, TRT introduces the token priority scoring module (TPSM). Then, we integrate the class activation map as the guidance to amend the produced context-aware feature maps. Extensive experiments on two benchmarks demonstrate that our proposed approach outperforms existing methods.

\section*{Acknowledgments}
This work is partially supported by the National Natural Science Foundation of China (Grant No. 62106235), by the Exploratory Research Project of Zhejiang Lab(2022PG0AN01), by the Zhejiang Provincial Natural Science Foundation of China (LQ21F020003), by Open Research Projects of Zhejiang Lab(NO. 2019KD0AD01/018).

\bibliography{egbib}
\end{document}